\documentclass[letterpaper, 10 pt, conference]{ieeeconf}  

\IEEEoverridecommandlockouts                              

\usepackage{amsmath,amsfonts}

\let\labelindent\relax
\usepackage{amssymb, amsthm}
\usepackage{mathtools}
\usepackage{dblfloatfix}

\usepackage{calc}
\usepackage{cases}
\usepackage{url}
\usepackage{hyperref}

\usepackage{float,color,graphicx}
\usepackage{tikz}
\usepackage{verbatim}
\usetikzlibrary{calc}
\usetikzlibrary{patterns}
\usetikzlibrary {arrows.meta}
\usepackage{pgfplots}
\pgfplotsset{compat=1.15}

\usepackage{enumitem}
\usepackage[font=footnotesize,skip=4pt]{caption}
\usepackage{cite}
\usepackage{balance}
\usepackage{tabularx}
\usepackage[position=t,singlelinecheck=off,caption=false]{subfig}
\usepackage{multirow}
\usepackage{enumerate}

\usepackage{mymath}
\usepackage{SLAM}
\usepackage{slam_methods}

\newcommand{\ignore}[1]{}

\title{\LARGE \bf
Task-driven SLAM Benchmarking for Robot Navigation
}

\author{{Yanwei Du$^{1}$, Shiyu Feng$^{2}$, Carlton G. Cort$^{1}$, Patricio A. Vela$^{1}$}
\thanks{*Supported in part by NSF Awards \#1816138, 2235944, \& 2345057.}
\thanks{$^{1}$Y.~Du, C.~G.~Cort, and P.~A.~Vela are with the 
School of Electrical and Computer Engineering, and 
Institute of Robotics and Intelligent Machines,  
Georgia Institute of Technology, Atlanta, GA 30308, USA.
{\tt\small \{yanwei.du, ccort6, pvela\}@gatech.edu}}
\thanks{$^{2}$S. Feng is with the School of Mechanical Engineering and the School of Electrical and Computer Engineering, Georgia Institute of Technology, Atlanta, GA 30308, USA.
{\tt\small shiyufeng@gatech.edu}}
}

\begin{document}

\maketitle
\thispagestyle{empty}
\pagestyle{empty}

\begin{abstract}

A critical use case of SLAM for mobile robots is to support localization during
task-directed navigation. 
Current SLAM benchmarks overlook the importance of repeatability (precision)
despite its impact on real-world deployments.
\textit{TaskSLAM-Bench}, a task-driven approach to SLAM benchmarking, addresses this gap.
It employs precision as a key metric, accounts for SLAM's mapping capabilities, and has
easy-to-meet requirements.
Simulated and real-world evaluation of SLAM methods provide insights into the navigation
performance of modern visual and LiDAR SLAM solutions. 
The outcomes show that passive stereo SLAM precision may match that of 2D LiDAR SLAM in indoor
environments.
%
\textit{TaskSLAM-Bench} complements existing benchmarks and offers richer assessment of SLAM 
performance in navigation-focused scenarios.
Publicly available code permits \textit{in-situ} SLAM testing in custom environments with
properly equipped robots.


\end{abstract}

\section{Introduction \label{sec:intro}}
Service robots have increasingly been deployed into home and work environments, and in public spaces
\cite{Barakeh2019PepperHR,dempseyAstro,tribelhorn2007evaluating,diligent2023,valdez2021humans,hotel2024,knightscope2016}.
Their primary goal is to provide repeatable services that enhance productivity and foster more efficient
work environments.
Central to the operation of these mobile robots is SLAM, whose localization capabilities support navigation.  
Robust and reliable SLAM is pivotal to completing tasks that are distributed throughout a given environment.
Despite recent advancements in SLAM achieving high accuracy in existing benchmarks and datasets 
\cite{Geiger2012CVPR, helmberger2022hilti, Burri2016euroc, schops2019bad}, service robots
using SLAM for feedback continue to face challenges with robustness such that localization
failures require human intervention to restore functionality \cite{liz2024wrangler}.
These closed-loop failures are not adequately addressed by existing open-loop SLAM
benchmarks, such that high performance in the former does not always translate to autonomous task-oriented 
performance.
For service robots, the main concern is reliable navigation to the same location when tasked,
rather than high accuracy absolute positioning. 
Additional perception-based modules ensure completion of the service activity near the goal
location.  
Consider that navigation benchmarks focus on robot terminal point performance, where success
is defined by the robot reaching the goal region \cite{Smith2019RealTimeEN, feng2021ego}.
A robot needs sufficient localization performance to reach the goal region.  
Of essence is reliable, repeatable performance to ensure successful task execution every time.
Repeatability, as measured using precision, is a common metric in robot automation
\cite{zhangEvalRepeatRobo, Paczek2018TestingOA, repeatInRobo, roboPreAcc}. 
This paper advocates to complement open-loop SLAM benchmarks with navigation-based SLAM benchmarks 
to better evaluate operational performance, and describes an easy to implement benchmarking
scheme.

\begin{figure}[t]
    \centering
    \includegraphics[width=0.48\textwidth]{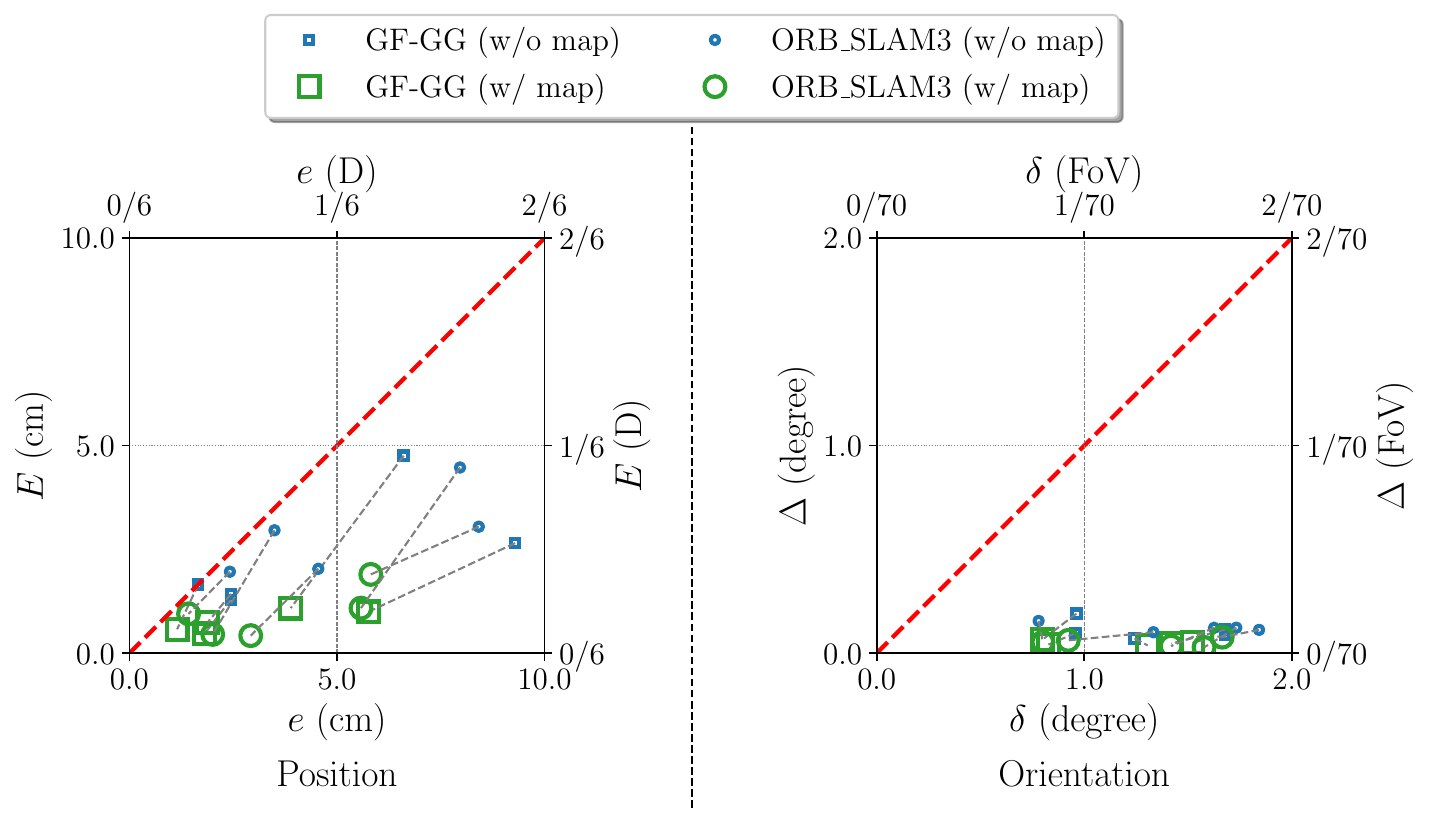}
    \vspace*{-1.6em}\caption{%
    Precision ($E$/$\Delta$) vs. Accuracy ($e$/$\delta$) in SLAM performance without (w/o) and 
    with (w/) a map for EuRoC Machine Hall sequences. Each marker is a single test under a specific
    test mode.  Dashed red lines mark precision equaling accuracy. Data points below
    the line suggest that accuracy under-reports task-repeatability.  Metrics and units are defined in
    \S\ref{sec:metric_defitions}, with experimental details given in \S\ref{sec:exp_ol_euroc}.%
    \label{fig:ol_euroc_pre_acc}}
    \vspace{-15pt}
\end{figure}

Real-world benchmarking with trajectory estimation error metrics \cite{grupp2017evo} remains mostly open-loop
due to metrics requiring absolute, global ground truth signals.  Significant effort is
required to obtain these signals. For instance, 
KITTI \cite{Geiger2012CVPR} uses RTK-GPS for precise vehicle positioning, and 
HILTI \cite{helmberger2022hilti} uses a total station (\textit{e.g.}, a survey grade laser 
scanner), both of which are costly. 
Datasets like EuRoC \cite{Burri2016euroc} and TUM \cite{sturm2012benchmark} use motion-capture 
systems, which are difficult to implement in large-scale indoor, multi-room environments. 
In these cases, methods like PennCOSYVIO \cite{penncosyvio} use
fiducial markers in conjunction with a laborious process involving meticulous placement
and measurement of the markers.  Sparse ceiling-mounted marker placement with total station
surveying and an extra upwards facing camera on the robot is effective for indoor settings
\cite{Kikkeri2014AnIM}.
Other methods \cite{kaveti2023challenges, wenzel2020fourseasons} rely on LiDAR-SLAM or
multi-sensor fusion results from \textit{an offline pose optimization process}
as ground truth. 
The proposed method removes the need for absolute measurement technology,
labor-involved annotation, and offline pose optimization.

The Subterranean (SubT) Challenge \cite{chuang2023into} was a real-world challenge 
requiring closed-loop implementation (\textit{e.g.}, SLAM and navigation integration) 
during exploration in underground environments with the objective of localizing placed targets. 
It prioritized target detection and target position in an absolute global frame.  
The evaluation criteria emphasized giving the true location of the target rather than being
able to reliably guide a human or robot back to the target. 
Resource intensive surveying methods generated the ground truth. 

While recent extensions have tackled rich, multi-sensor fusion capable benchmarking
\cite{kaveti2023challenges, helmberger2022hilti}, the performance gap between SLAM benchmarking via
open-loop, sensor stream replay versus task-oriented performance motivates others to evaluate SLAM and
related autonomy modules through different lenses. Simulated worlds figure prominently due to
reproducibility of outcomes, ease of implementation by researchers, and the availability of ground truth
information \cite{zhao2020closed}.
Embodied AI research, such as the Habitat Navigation Challenge \cite{habitatchallenge2023}, exemplifies
the necessity for a task-driven evaluation methodology.  The simulation-based challenge  emphasizes the
need for reliable, long-term navigation capabilities.
The importance of localization for autonomous vehicles motivated a simulation-based benchmarking scheme for
street navigation \cite{suomela2023benchmarking} with success/failure as the main metric. 
%
SLAM latency plays a role in its robustness and accuracy, with a relationship found between these two
factors for trajectory tracking tasks using SLAM pose estimates as feedback \cite{zhao2020closed, zhao2020good}.
Simulation-based benchmarking \cite{zhao2020closed, zhao2020good} provides support for moving
beyond accuracy when considering SLAM in the closed-loop. A reasonable next step would more
explicitly consider navigation.

An additional difference between these benchmarks and service robot deployment is that the latter often
incorporates an initial mapping stage.
Existing benchmarks overlook this phase by evaluating performance solely through one-time replay. 
SLAM maps mitigate long-term drift and ensure consistency (high precision) across
repeat visits to the same locations \cite{campos2021orb,mur2017visual}.
Multi-session SLAM evaluation \cite{pearson2023robust} does consider map reuse with an accuracy metric for repeated runs to the same locations. 
However, the study design and variables may consider different performance factors and experimental
methodology for LiDAR SLAM \cite{pearson2023robust}, and also for visual SLAM multi-map and map reuse
open-loop tests \cite{campos2021orb,mur2017visual}. These works point to the need for reproducible and
standardized benchmarking schemes for SLAM systems equipped with pre-built maps.  Further evidence for
this need is given in Fig.~\ref{fig:ol_euroc_pre_acc}, with multi-run testing on the EuRoC dataset (see \S\ref{sec:exp_ol_euroc} for details).  
The Precision vs Accuracy plots show better precision than accuracy indicating that SLAM evaluation may under-report performance.  Including a mapping phase further improves performance, especially position precision.  
Referencing the performance metrics to robot characteristics (under $x$-axis) permits qualitative assessment of 
potential goal attainment success by relating precision to the visible task region at the robot's arrival
pose. 

This paper introduces \textit{TaskSLAM-Bench}, a navigation-in-the-loop SLAM benchmark framework
emphasizing task-relevant metrics: repeatability (precision) and completion. 
The benchmarking approach mimics real-world implementations where robot deployment may include an
autonomous map building phase for improved task execution during deployment.
Experiments with visual and LiDAR SLAM methods identify which methods effectively support navigation tasks. 
The key contributions are:
\begin{itemize}[leftmargin=*]
    \item A low-cost, easy to setup, task-driven SLAM benchmark with precision 
    as the key metric for measuring robot consistency in reaching the same pose over
    multiple rounds. The method scales to large environments.
    \item Evaluation of stereo visual and LiDAR SLAM systems in simulation and real-world
    scenarios, with evidence that passive visual methods can match the robustness and
    precision of LiDAR-based methods for indoor settings.
    \item Open-sourced benchmarking code \cite{taskDrivenSlamBench} for the robot
    autonomy community to easily test and evaluate task-driven SLAM in their own
    environments \& on their own robots.
\end{itemize}

\ignore{
For SLAM map reuse, a similar approach, known as multi-session SLAM, has been explored in works such as \cite{campos2021orb} and \cite{mur2017visual}. However, a key distinction in our testing is that they use the first sequence for mapping and process subsequent sequences with the map, whereas we treat each sequence independently. In our case, the SLAM system builds the map in the first round and re-uses the map in subsequent rounds within that sequence. Our aim is to avoid the mapping coverage limitations inherent in restricted sequences and instead maximize the utility of SLAM maps throughout testing. This is a practical approach that mimics real-world implementations where robots are typically given ample time to create a complete map, ensuring task execution.}


\section{Methodology \label{sec:meth}}
\subsection{Preliminaries}

The following symbols and definitions are used:

\begin{itemize}[leftmargin=*]
  \item \textbf{Robot Pose} - \((x, y, \theta)\) - Robot position and orientation. $SE(3)$ pose estimates are projected to $SE(2)$.

  \item \textbf{Sequence} - \(\mathbf{S}\) -  A set of waypoints the robot is required to sequentially follow, i.e., \(\mathbf{S} = \{\mathbf{w}_1, \mathbf{w}_2, \dots, \mathbf{w}_N\}\).

  \item \textbf{Waypoint} - \(\mathbf{w}_i = (x_i, y_i, \theta_i)\) indexed by $i$ - position and orientation (heading angle) of the robot in the 2D plane.

  \item \textbf{Indexing} - $i$ represents the waypoint index, and $j$ the round index, and 
    $k$ the sequence index. 
  
  \item \textbf{Counts} - $N$ is the no. of waypoints in a sequence, $M$ is the no. of 
    rounds in a test, and $K$ is sequence count. 
\end{itemize}
Variables with a {$^*$ symbol indicate ground truth or reference values}, while a $\,\Bar{}\,$ symbol denotes mean value.

\subsection{Task-Driven SLAM Benchmarking Design}

\begin{figure}[t]
  \vspace{0.06in}
  \centering
  \includegraphics[width=0.9\linewidth]{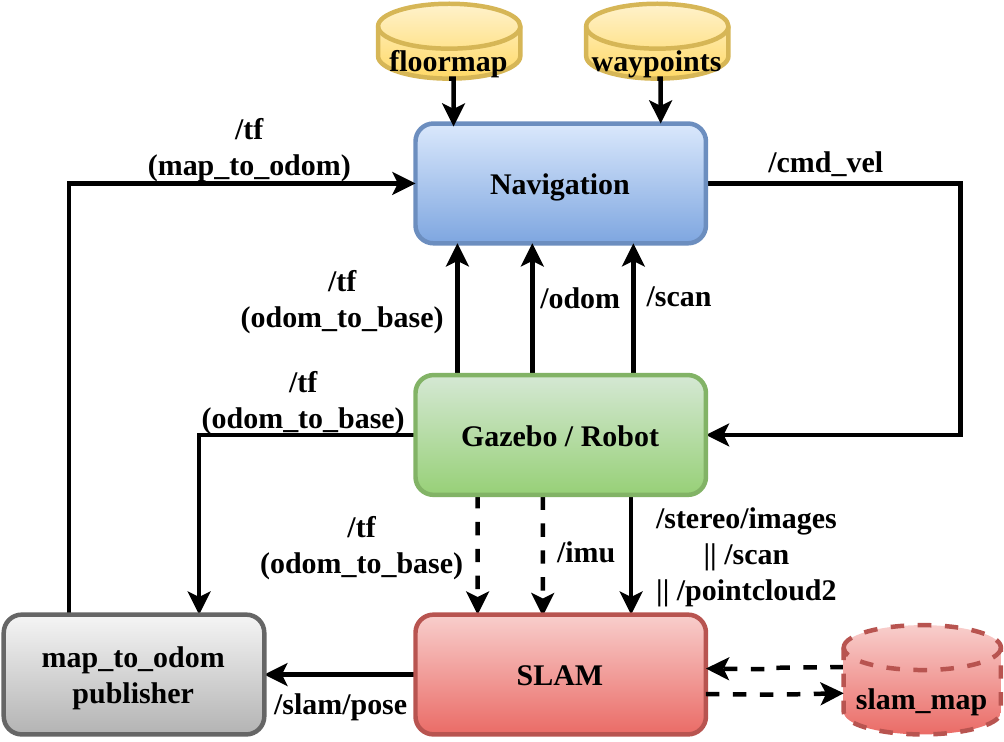}
  \caption{Task-driven SLAM benchmarking design. Dashed arrows and cylinders represent optional 
    elements specific to SLAM methods and test configurations. The $||$ symbol denotes 
    alternative robot sensor configurations.}
  \label{fig:framework}
  \vspace*{8pt}
  \includegraphics[trim=1cm -1cm 0cm 0cm, width=0.45\textwidth]{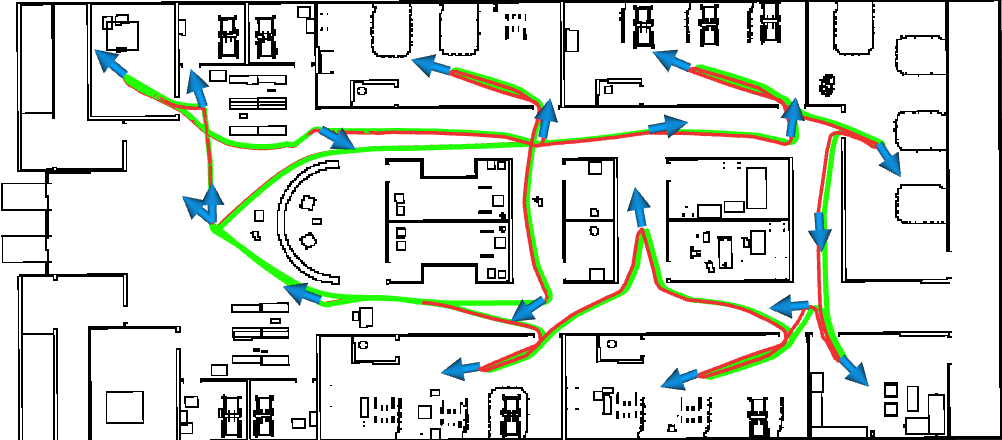}
  \vspace*{-4pt}
  \caption{Defined waypoints (blue arrows) in the AWS Gazebo hospital world, with 
    the robot estimated trajectory (red) and ground truth trajectory (green) plotted when
    executing the navigation task.%
    }\label{fig:task_def} \label{fig:aws_hospital_wpts}
  \vspace{-15pt}
\end{figure}

Mobile robot task execution requires SLAM to support collision-free navigation to designated
target locations.  Hence, the proposed SLAM benchmarking framework evaluates the use of
closed-loop SLAM operation as part of navigation.
Fig.~\ref{fig:framework} depicts the navigation system as implemented in ROS
\cite{quigley2009ros}. The navigation module uses TEB \cite{rosmann2013teb}.

\subsubsection{Task Definition}
Waypoint Navigation. 
Waypoints represent task-relevant goal points, as exemplified in the AWS Gazebo hospital model 
\cite{awsgazebomodel} of Fig.~\ref{fig:task_def}.
They are manually specified based on a floor plan of the environment's free space. 
It is assumed that the floor plan reflects the static environment, with no dynamic obstacles or
unknown areas.
The robot starts at a fixed location and sequentially navigates to each waypoint. 
Upon reaching a waypoint, the robot's pose is recorded for evaluation.
The procedure repeats for multiple rounds, terminating if the robot fails to reach a
waypoint. 

\subsubsection{Performance Metrics} \label{sec:metric_defitions} 
Accuracy, Completeness, and Precision. 
Accuracy metrics such as Absolute Trajectory Error (ATE) or Relative Pose Error (RPE)
typically measure SLAM performance \cite{sturm2012benchmark}. 
Navigation and trajectory tracking evaluation of SLAM employ task-specific metrics \cite{suomela2023benchmarking,zhao2020closed}.
Here, evaluation of indoor mobile robot navigation emphasizes waypoint navigation 
\textit{reliability and repeatability}, measured by successful goal attainment percentage (completeness) and goal pose precision.
Simulated scenarios include accuracy, in the form of APE, since ground-truth is
available.

Navigation \textit{accuracy} to waypoint $\textbf{w}_i$ captures how close the robot gets to the
target goal over repeated rounds. It is based on the average position and orientation errors:
\begin{equation} \label{eq:wpt_acc}
    e_{\textbf{w}_i} = \frac{1}{M}\sum_j^M \lVert \mathbf{z}_{ij} - {\mathbf{z}}_i^* \rVert_2 
    \ \ \text{and} \ \ 
    \delta_{\textbf{w}_i} = \frac{1}{M}\sum_j^M \lVert \mathbf{\theta}_{ij} -
    {\mathbf{\theta}}_i^* \rVert,
\end{equation}
where $\mathbf{z} = (x, y)$ and orientation error factors for wrap-around.
Navigation \textit{precision} at waypoint $\textbf{w}_i$ quantifies the proximity of robot
pose measurements to each other over repeated rounds. Decomposed into position and
orientation:
\begin{equation} \label{eq:wpt_pre}
    E_{\textbf{w}_i} = \frac{1}{M}\sum_{j}^{M} \lVert \mathbf{z}_{ij} - \Bar{\mathbf{z}}_i \rVert_2
    \ \text{and} \ 
    \Delta_{\textbf{w}_1} = \frac{1}{M}\sum_j^M \lVert \mathbf{\theta}_{ij} -
    \Bar{\mathbf{\theta}}_i \rVert,
\end{equation}
where means are computed over all rounds (index $j$). 

Completeness ($C$) refers to the ratio of completed waypoints over all waypoints across all sequences:
\begin{equation} \label{eq:seq_completeness}
    C = \frac{\sum_{k=1}^{K}\sum_{i=1}^{N_k} \delta_{ki}}{\sum_{k=1}^{K} N_k}
    \ \ \text{for} \ \ 
    \delta_{ki} =
    \begin{cases}
        0, & \text{ if } M_{ki} < M \\
        1, & \text{otherwise} 
    \end{cases}
\end{equation}
where $N_k$ is the waypoint count in sequence $S_k$, and $M_{ki}$ is the successful waypoint 
$\textbf{w}_i$ attainment count for the $M$ runs.

\textbf{Task-driven Units} - $(D, FoV)$. 
To emphasize a task-driven perspective, outcomes will also report in unit scales 
related to robot specific properties: \(D\) the robot's diameter, and \(FoV\) the camera's horizontal field of view. 

\subsubsection{Measuring Completeness and Precision} \label{sec:realworld_performance_measuring}
Since the task-driven benchmark prioritizes waypoint precision, which assesses the
repeatability of the robot’s navigation across multiple rounds relative to its own internal
(SLAM) coordinate system,
there is no need to establish a global absolute coordinate system. Precision at each
waypoint will be recoverd from individual reference frames.

\begin{figure}[t]
  \vspace*{0.06in}
  \centering
  \begin{tikzpicture}[inner sep=0pt,outer sep=0pt]
    \node[anchor=north west] (tag) at ($(0, 0)$)
      {{\includegraphics[width=0.80in]{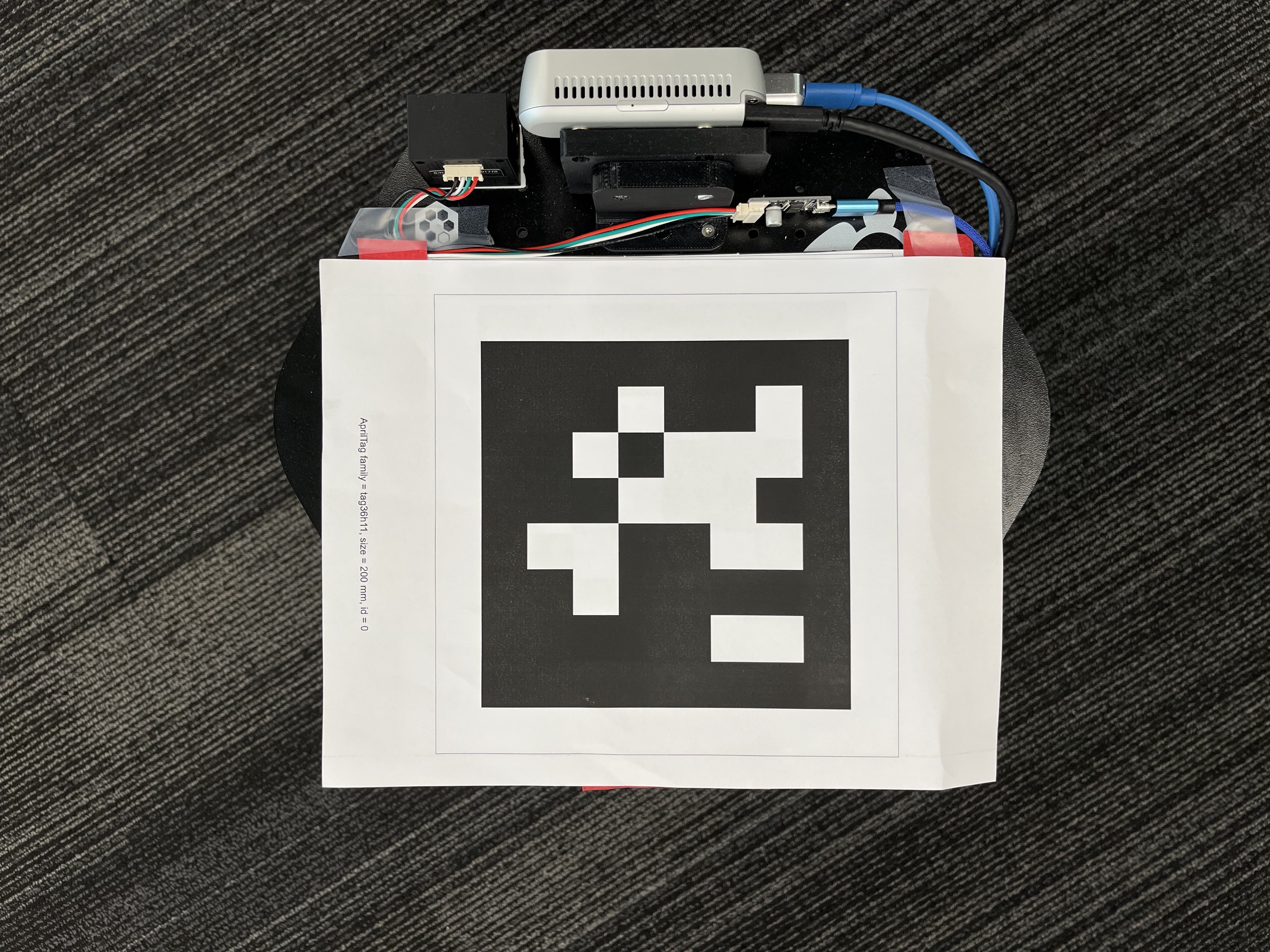}}};
    \node[anchor=north,yshift=-3pt] (camera) at ($(tag.south)$)
      {{\includegraphics[width=0.80in,clip=true,trim=0.1in 0.1in 0.1in 0.1in]{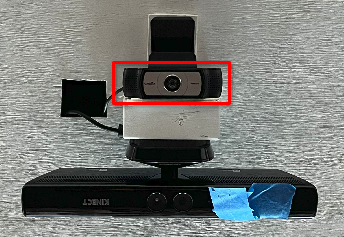}}};
    \node[anchor=north east, xshift=-6pt] (turtlebot) at ($(tag.north west)$)
      {{\includegraphics[width=0.28\columnwidth]{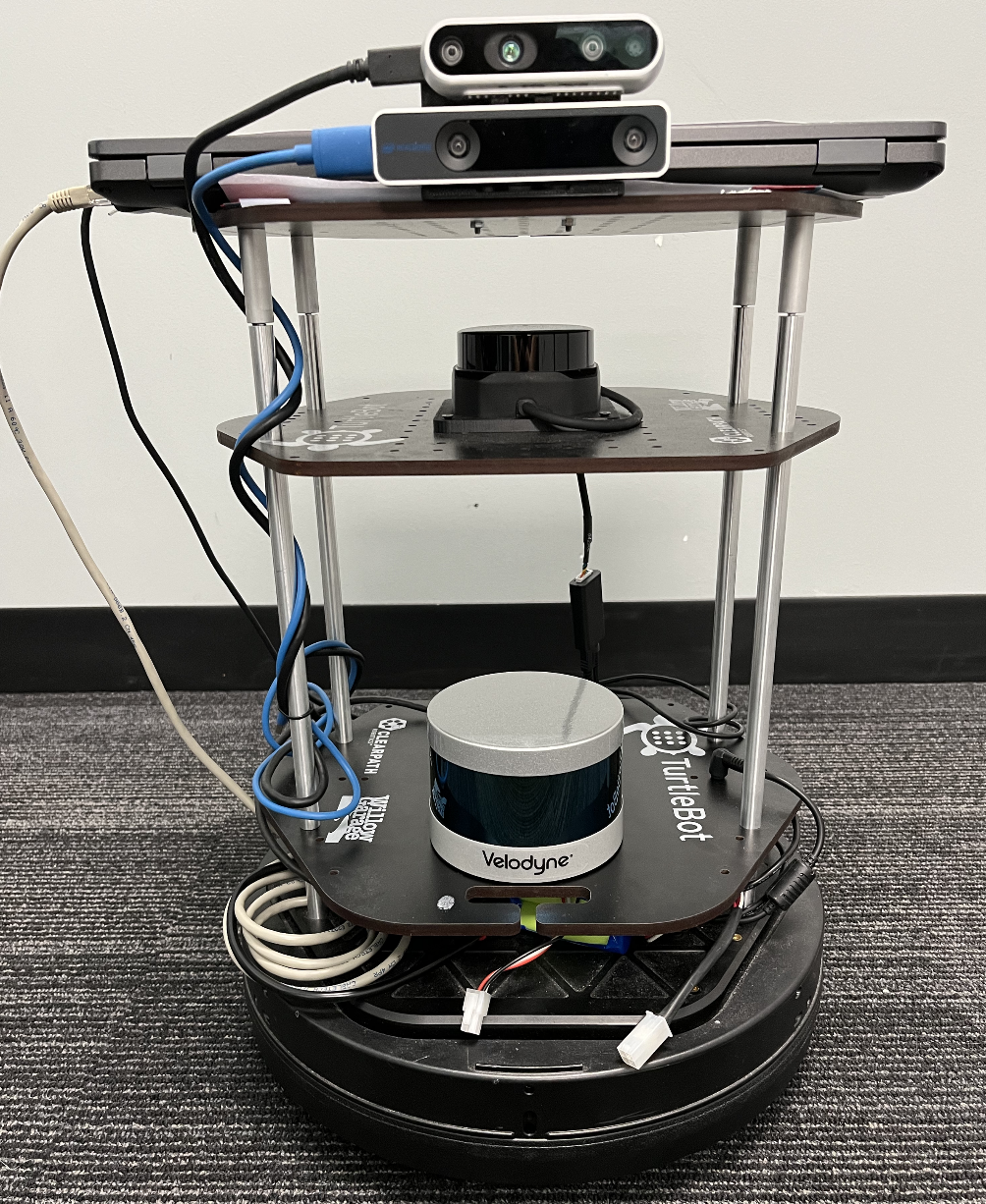}}};
    \node[anchor=north west] (tsrb) at ($(tag.north east)+(0.5em,-0.8em)$)
      {{\includegraphics[height=1.05in,clip=true,trim=0.4in 0.0in 0.175in 0.5in]{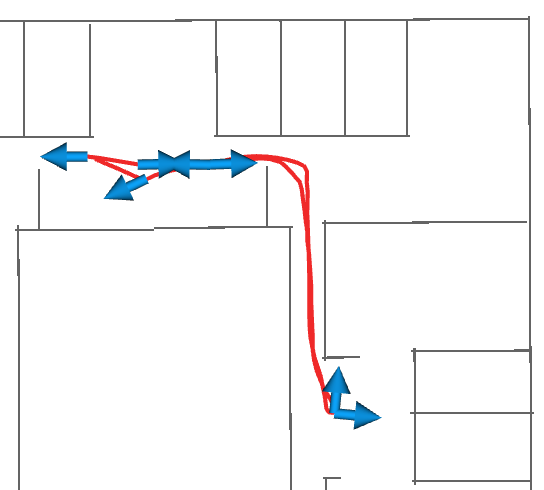}}};
    \node[anchor=north east] (real_wpts) at (tsrb.south east)
      {{\includegraphics[height=0.650in,clip=true,trim=0.05in 1.9in 3.4in 0.98in]{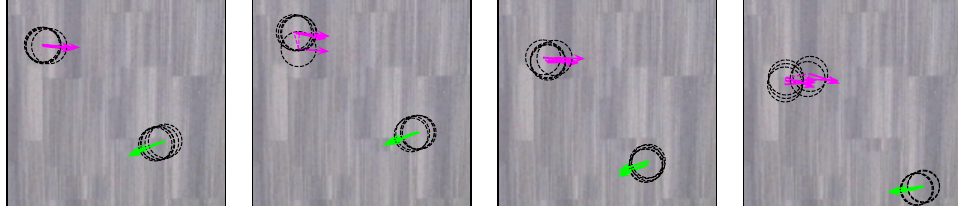}}}
        node[above] at ($(real_wpts.south west)+(2.0em,0.1em)$) {\tiny \textbf{{\orb}}}
        node[above,xshift=50pt] at ($(real_wpts.south west)+(2.0em,0.1em)$) {\tiny \textbf{{\gfgg}}}
        node[above,xshift=112pt] at ($(real_wpts.south west)+(2.0em,0.1em)$) {\tiny \textbf{{\slamtoolbox}}}
        node[above,xshift=165pt] at ($(real_wpts.south west)+(2.0em,0.1em)$) {\tiny \textbf{{\fastlio}}};
  \end{tikzpicture}
  \caption{%
  Real-world experimental setup. Clock-wise from top-left:  
  A Turtlebot2 equipped with SLAM sensors and laptop, on top of which is placed an AprilTag.
  Downward facing cameras (in red box) and computers/laptops placed in the environment detect and recover
  the robot's pose.  
  The robot waypoints, defined to be under the cameras, should be accurate enough that the robot will be
  visible when reached. During the programmed tour, the cameras will detect and estimate the robot pose
  when seen. Precision and Completeness follows from the estimates. 
  The lower figure presents an example of the robot captured at two waypoints (magenta and
  green) for different SLAM methods. Arrows indicate orientation and black circles indicates
  position, over multiple runs.%
  \label{fig:overhead_cam}\label{fig:apriltag}\label{fig:turtlebot}\label{fig:tsrb_map_wpts}}
  \vspace*{-1em}
\end{figure}

Waypoint precision measurement in real-world conditions may be obtained in two equivalent
manners similar to virtual reality systems: outside-in or inside-out. Outside-in uses a
downward facing camera from a suitable height (Fig. \ref{fig:overhead_cam}). 
Options include overhead/ceiling-mounted, tripod-installed, or otherwise placed to face the
ground region the robot should reach.
Each camera requires connection to a (low cost) computer or laptop to estimate the robot’s
pose upon waypoint attainment, based on an AprilTag affixed to the top of the robot 
(Fig. \ref{fig:apriltag}).  
Inside-out operates as in
\cite{Kikkeri2014AnIM} by inverting the setup: an upwards-facing camera is mounted on the
robot with tags placed in the environment.
AprilTags work for SLAM benchmarking \cite{penncosyvio}, 
with the C++ detection library achieving sub-pixel accuracy ($\leq$ 0.5)\cite{Kallwies2020DeterminingAI}. 
Prior study \cite{abbas2019analysis}\cite{Kalaitzakis2020ExperimentalCO} showed acceptable pose
estimation accuracy for distances and viewpoints compatible with our indoor setup.

Importantly, waypoint precision does not rely on absolute references. 
It is assessed independently of other waypoints, with evaluation done in local
coordinate systems defined by the camera and tag frames.  
The outside-in approach does not require global calibration of the camera poses (nor
relative poses).  Each camera/waypoint measurement pair operates independently.  The
inside-out approach approach does not require a global scan nor globally aligned tags. 
Waypoint attainment success is established by the tag being fully visible to the
camera, otherwise waypoint attainment failed. 

From a task-driven perspective, as long as the robot can successfully transit to subsequent tasks, 
achieving high accuracy is less critical.  Instead, high precision indicates that the robot
consistently navigates to the target location in its internal SLAM coordinate system, which is
better aligned with the needs of practical applications.
This method offers a significant testing advantage.  Both setups are straightforward and
easy to install at the necessary measurement areas.  They do not require global calibration,
specialized expertise, nor expensive specialized equipment for operation
(as required for a motion capture system or a surveying strategy).
Since simulation experiments provide accuracy and precision from simulator ground truth
information, they provide understanding of how these two metrics relate in a given
environment.

%


\subsubsection{Map-Based Performance Evaluation}\label{meth:map-based-eval} To demonstrate the impact of map reuse on task completion, we run additional tests for SLAM methods that support mapping. Since some methods cannot save and load maps, we introduce an extra priming round where the robot traverses the waypoints and builds a map online. This map is then used for subsequent rounds without resetting the robot's pose to the initial location after each round. 
The robot poses recorded during the mapping phase are excluded from evaluation.
Comparing the results from tests with and without the use of a pre-built SLAM map highlights the benefits of a prior SLAM map for task-oriented SLAM performance.

\section{Experiment Design \label{sec:exp}}
Effectiveness of the \textit{TaskSLAM-Bench} framework will be established through a series of controlled 
experiments. 
Simulation experiments assess SLAM performance using all metrics: Accuracy, Precision, and Completeness.
%
Plots will include dual axes with standard units and robot-relative units to link
performance directly to robot characteristics and task requirements.
If a robot is too far from an object, relative to its size, or headed in a bad direction,
relative to its visual field of view, then a follow-up visually guided manipulation,
inspection/object viewing, or other task would fail.

\subsection{SLAM Candidates}\label{subsec:slam_candidates}
To demonstrate the benchmark flexibility and to perform comparative analysis, several SLAM 
methods are tested. They include inertial and robot odometry signals when required or permitted.
SLAM parameters were not tuned for optimal performance across the tests, with default settings used;
they are generally effective in most scenarios and offer balanced performance. 
Our goal is not to identify the best SLAM method, but rather to demonstrate the functionality of 
the proposed SLAM benchmarking design (\textit{TaskSLAM-Bench}), and to use the
demonstration to assess which SLAM methods are suitable for task-driven purposes. Tested are:

\subsubsection{2D LiDAR}
{\slamtoolbox}\cite{Macenski2021} and  {\hectorslam}\cite{kohlbrecher2011flexible}.
They leverage scan matching to correct robot odometry drift and are commonly used for indoor navigation.

\subsubsection{Stereo Visual and Visual-Inertial}
{\gfgg}\cite{zhao2020good} \& {\orb}\cite{campos2021orb}, feature-based stereo visual methods;
{\dsol}\cite{qu2022dsol}, a direct odometry system;
{\svo}\cite{forster2016svo}, a semi-direct method; and 
{\msckf}\cite{mourikis2007multi}, a filter-based method.
{\orb} is run stereo-only as it would fail to initialize with inertial sensors, even with extra
motion before testing. 

\subsubsection{3D LiDAR-Inertial}
{\fastlio}\cite{xu2021fast} and
{\liosam}\cite{shan2020lio}.
They are commonly deployed for autonomous driving, and were used by SubT Challenge competitors \cite{Ebadi2022PresentAF, koval2022evaluation}.

\begin{table}[t]
  \vspace*{0.06in}
\centering
\caption{Scenario Properties in Simulation and Real-World}\label{tab:cl_scenarios}
  \begin{tabular}{| c | c | c | c |}
  \hline 
  \textbf{Name} & \textbf{Key Features} & \begin{tabular}{@{}c@{}} \textbf{Area} \\ \textbf{$(m^2)$} \end{tabular} & \begin{tabular}{@{}c@{}} \textbf{Path} \\ \textbf{Length} \\ \textbf{$(m)$} \end{tabular}\\ \hline
Small House & Home furniture layouts & 144 & 45 \\ \hline
Warehouse & Shelves with boxes \& goods & 260 & 70 \\ \hline
Hospital & Rooms w/medical equipment & 1400 & 220 \\ \hline
TSRB Office & Multiple long corridors & 1500 & 260 \\ \hline \hline
\begin{tabular}{@{}c@{}} TSRB Short \end{tabular} & A long corridor and sharp turns & 225 & 45 \\ \hline 
\begin{tabular}{@{}c@{}} TSRB Long  \end{tabular} & Long distance loop closures &
620 & 120 \\ \hline
\end{tabular}
\vspace*{-2em}
\end{table}

\subsection{Closed-Loop-SLAM Navigation Experiments}
Testing involves both simulated and real-world environments, whose scenarios are described in
Table \ref{tab:cl_scenarios}.  In each scenario, the robot starts from a fixed initial
location and navigates sequentially through the waypoints.
The navigation system configuration is fixed across all experiments.
Upon reaching each waypoint, the robot pauses for 5 seconds before proceeding to the next one. 
For an outside-in setup, the computer/laptop at this waypoint, synchronized with NTP (delay $<$ 1 second), saves timestamped image data upon detecting the tag affixed to the robot. 
The robot's poses are estimated offline and associated with waypoints by timestamp for
performance evaluation.

A single scenario is tested for five consecutive rounds. In each round, the robot is either
reset or continues based on the testing modes—without and with a SLAM map. 
The same resetting strategy is applied to all SLAM methods tested. 
A waypoint is considered successfully completed only if the robot arrives at it in all five
rounds; failure to reach a waypoint in any round leads to the incomplete classification.
Accuracy, if available, and precision for each completed waypoint is
calculated according to Eqns. (\ref{eq:wpt_acc}) and (\ref{eq:wpt_pre}).

The testing platform is a Turtlebot 2 robot, shown in Fig.~\ref{fig:turtlebot}, which offers
odometry measurements through built-in sensors. 
Mounted to it are an RPLiDAR S2 for 2D LiDAR-based SLAM methods, 
a Realsense D435i (FoV: 87$^\circ$) for visual (and visual-inertial) methods, and 
a Velodyne-16 for 3D LiDAR SLAM methods.  The laser data also feeds to the navigation module
for obstacle avoidance.  All the processes run on an Intel Core i7-9850 laptop
(single-thread passmark score of 2483).  The simulation robot is similarly configured.
The task-driven units use \(D = 30\) cm and \(FoV = 80.0^{\circ}\). 

\subsubsection{Simulation}
Gazebo\cite{koenig2004gazebo} is chosen for the simulation benchmark testing. It provides
readily available sensor setups (\textit{i.e.}, camera and LiDAR feeds) and ground-truth data
for measuring accuracy and precision. 

\subsubsection{Real World}
Real-world testing was performed in an office environment. One test defined a path sequence 
of $\sim\!45$ meters in length and featuring six waypoints, see Fig.~\ref{fig:tsrb_map_wpts}. 
It has short stretches in a corridor, sharp turns, and return loops.
Using the outside-in approach, three overhead cameras (Fig. \ref{fig:overhead_cam}) captured
the robot for success determination and precision measurement.  
Another test, with the inside-out approach, defined a travel sequence of about 120 meters in
length with twelve waypoints.  It has similar properties to the first test.
Real-world sequences include nuisace factors like sensor noise, hardware delays, 
and low-texture areas.
\begin{figure}[t]
  \vspace*{0.06in}
  \centering
  \begin{tikzpicture}[inner sep=0pt, outer sep=0pt]
    \node[anchor=north west] (sim_pre_acc) at ($(0, 0)$)
        {{\includegraphics[trim=0cm 0 0cm 0cm, width=0.47\textwidth]{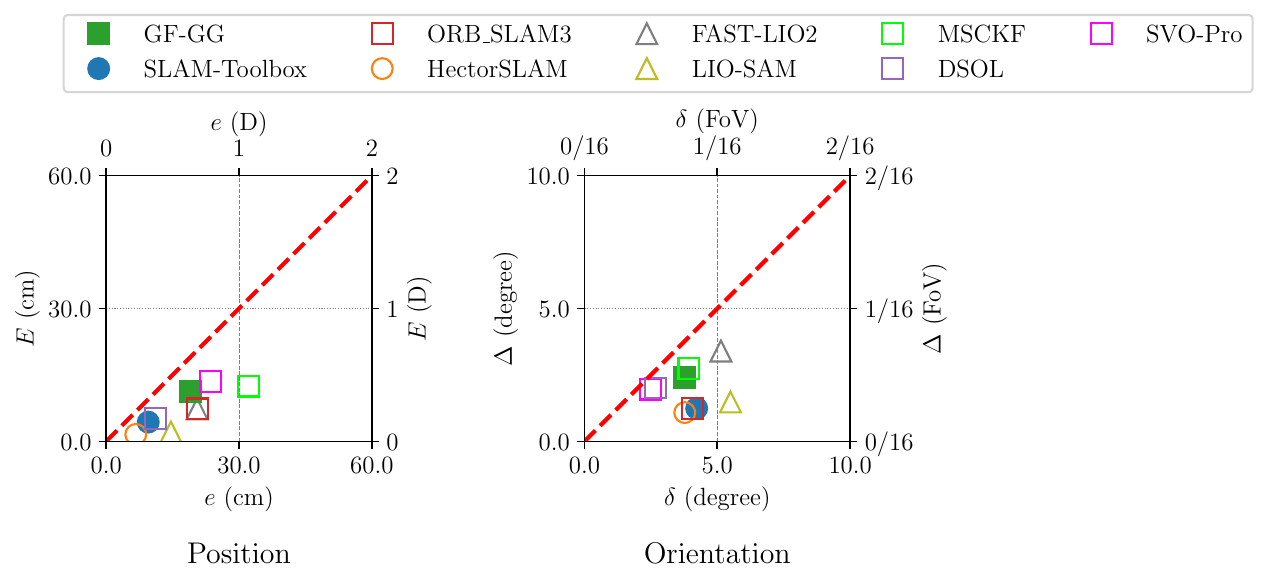}}};

    \node[anchor=north west] (legend) at ($(sim_pre_acc.east)+(-5.2em,2.5em)$)
        {{\includegraphics[trim=0cm 0 0 0cm, width=0.13\textwidth]{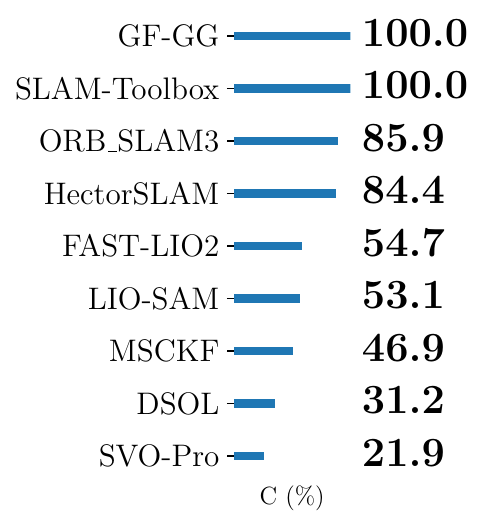}}};

    \draw [dashed] ($(sim_pre_acc.north)+(-3.3em, -1.8em)$) -- ($(sim_pre_acc.south)+(-3.3em, 0.8em)$);
    \draw [dashed] ($(sim_pre_acc.north east)+(-5.3em, -1.8em)$) -- ($(sim_pre_acc.south east)+(-5.3em, 0.8em)$);

  \end{tikzpicture}
  \caption{
    Simulation SLAM Precision vs. Accuracy in tests w/o map.
    Methods with \(100\%\) waypoint completeness have filled markers.
    The dashed red line denotes precision equal to accuracy. 
    Markers below this line mean that accuracy under-reports task-repeatability performance.
    Completeness values are represented as a bar plot on the right (longer is better).%
    \label{fig:sim_pre_acc}}
    \vspace{-12pt}
\end{figure}

\subsection{Open-Loop EuRoC Evaluation - A Preliminary Study} \label{sec:exp_ol_euroc}

This section describes the performance evaluation performed for the provisional SLAM
performance analysis noted in the Introduction, which argued that performance
evaluation would benefit from including task-driven evaluation. 
The commonly used EuRoC machine hall sequences ground this preliminary study to other SLAM benchmarking works. 
Given the lack of distinct waypoints, precision and accuracy are based on the pose estimates at each
frame.
%

\subsubsection{Setup}
Two stereo SLAM methods, {\orb} and {\gfgg}, are tested. 
For \textit{w/map}, each sequence restarts without resetting the SLAM method to keep the map in memory for subsequent runs. 
Eqns.~(\ref{eq:wpt_acc}) and (\ref{eq:wpt_pre}) are modified for evaluation in $SE(3)$. 
Based on \cite{Burri2016euroc}, task-driven units are set to $D = 30 cm$ and $FoV =
70.0^{\circ}$. 

\subsubsection{Results and Analysis}
Fig. \ref{fig:ol_euroc_pre_acc} visualizes the results as Precision vs. Accuracy plots.
Given that the red dashed line is precision-accuracy parity, position precision values match
or are lower than accuracy values across all sequences.  Orientation precision is noticeably
better than accuracy. The results exhibit better localization repeatability than absolute
accuracy would suggest.  
When utilizing a SLAM map, both methods exhibit improved performance with outcomes shifted
toward the lower-left corner.  Position precision error, which is within $1/6$ of a robot
diameter in \textit{w/o map} tests, reduces to $1/12$ of a robot diameter in \textit{w/map} tests.
Orientation precision error, at $1/35\,FoV$, reduces by a factor of $10$ in the \textit{w/map} tests
(to pixel level error).  An object intended to be centered in the field of view from a
reasonable distance would still be visible and nearly centered.  This suggests that accuracy
does not fully reflect potential SLAM performance from a task-driven perspective, and points
towards the benefits of \textit{augmenting} it with precision.

\subsection{Task-Driven SLAM Benchmark Outcomes and Analysis}
\begin{figure}[t]
  \vspace*{0.06in}
  \centering
  \begin{tikzpicture}[inner sep=0pt, outer sep=0pt]
    \node[anchor=north west] (sim_pre_cum) at ($(0, 0)$)
        {{\includegraphics[trim=0cm 0 0cm 0cm, width=0.45\textwidth]{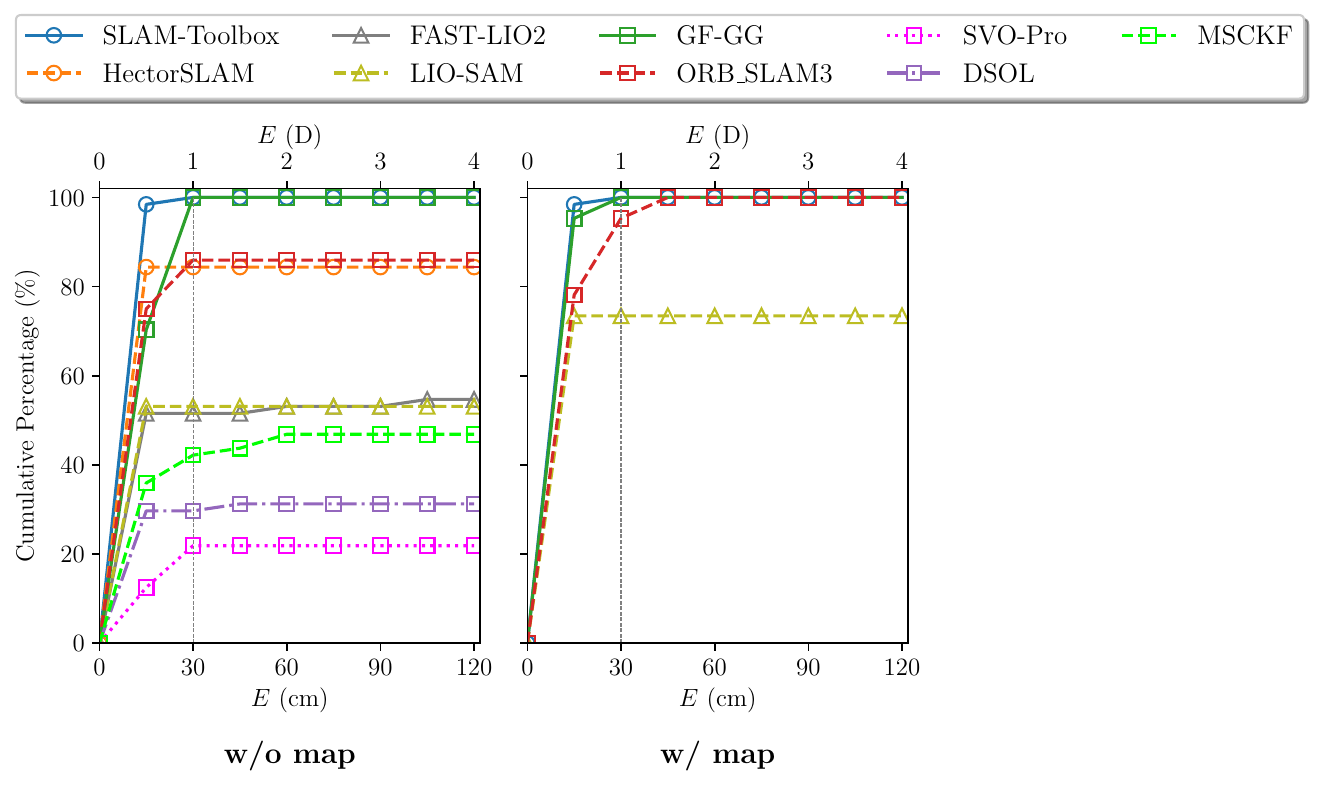}}};

    \node[anchor=north west] (legend) at ($(sim_pre_cum.east)+(-6.5em,4.5em)$)
        {{\includegraphics[trim=0cm 0 0 0cm, width=0.12\textwidth]{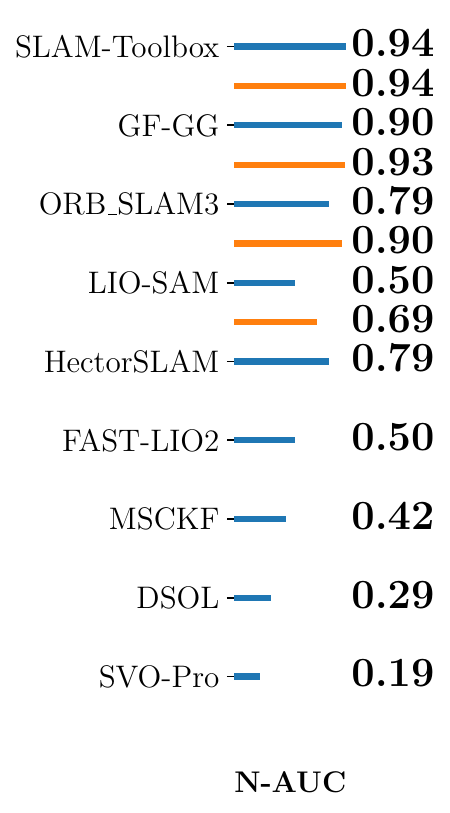}}};

  \end{tikzpicture}
  \caption{%
    Cumulative Position Precision plots for tests w/o-and w/-map in simulation.
    Normalized area-under-curve for each method in both modes, shown as a bar plot on the right. 
    Blue lines: w/o-map; orange lines: w/-map. Longer bars indicate better performance.%
    \label{fig:sim_wpt_pre_cum}}
    \vspace{-15pt}
\end{figure}

Simulation outcomes will first be analyzed, including precision/accuracy comparison. 
Real-world outcomes will then be analyzed.  Following \cite{qu2022dsol}, performance
analysis will include cumulative precision plots (see Figs.~\ref{fig:sim_wpt_pre_cum} and
\ref{fig:realworld_pre_cum}). 
The curve traces the percentage of waypoints with a precision error value less than the
equivalent $x$-axis threshold.  
The normalized area-under-curve (N-AUC) for the cumulative plots provides an overall
performance score for each method.  A higher N-AUC indicates better performance, with curves
closer to the top-left corner representing superior precision and completeness.
The analysis points to precision as a meaningful metric to pair with completeness, and
to passive stereo SLAM as being on par with LiDAR-based SLAM.

\subsubsection{Simulation}
Fig.~\ref{fig:sim_pre_acc} Precision vs Accuracy plots in without map tests consistently show lower
precision value than accuracy when the navigation succeeds. To some degree this indicates that
localization failure is catastrophic when it occurs; there doesn't seem to be a sliding scale. The
cumulative position precision plot in
Fig.~\ref{fig:sim_wpt_pre_cum} indicates the same by plateauing rather than exhibiting
linear growth to 100\% across the precision axis. In the cases studied, either a SLAM method
is precise to a robot diameter or it fails. 
Orientation accuracy and precision values are low given the field of view. Due to the strong
performance of orientation, it will not be reviewed in future analyses.

The least complete implementations are odometry methods ({\dsol}, {\svo}), and {\msckf} 
as they cannot leverage map content, nor long baseline associations outside of the filtering
window. The two 3D-LiDAR methods performed comparably in the w/o map tests. 
However, they did not match 2D-LiDAR performance.
This performance contrasts with their near universal preference in robot deployments
\cite{chuang2023into}, which may be due to the heavy emphasis on \textit{exploration}
rather than repeatable navigation. Exploration has looser navigation demands.
{\hectorslam} and {\orb} were closer to perfect, but exhibit enough failures to not be reliable.
One visual SLAM method, {\gfgg}, matched the perfect completeness performance of the
2D-LiDAR {\slamtoolbox} implementation. Both achieved 100\% cumulative precision at one
robot diameter, indicating that the two methods are comparable in performance, though
{\slamtoolbox} has an edge in the N-AUC score.  


Using a map improves performance as seen in Fig.~\ref{fig:sim_wpt_pre_cum} by the upwards
and/or leftwards shift of plotted curves relative to the w/o map curve. 
Correspondingly, the N-AUC scores improve.  
{\slamtoolbox} shows minimal change, primarily due to its scan-to-map registration design,
which is well-suited for indoor scenarios, and its continuous pose-graph-based global map
management.
{\orb}'s imperfect w/o map completion becomes perfect w/map completion. It has a lower
N-AUC score relative to {\slamtoolbox} and {\gfgg}, hitting perfect completion at around
$1.5D$.

\begin{figure}[t]
    \centering
    \includegraphics[width=0.45\textwidth,clip=true,trim=0in 0in 0in 2in]{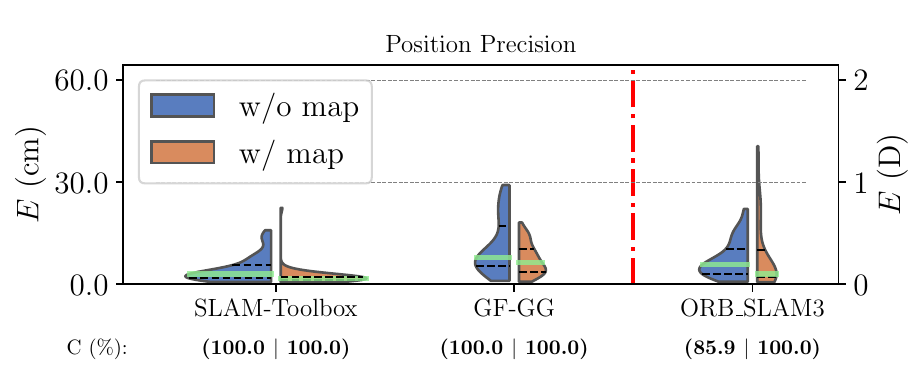}
    \caption{
    Distribution of Waypoint Position Precision for tests without (w/o) and with (w/) map in
    simulation.
    The vertical red dash-dotted line marks the boundary for full completeness.
    Each distribution bar contains the mean (solid green line) and quartiles (black dashed lines).}
    \label{fig:sim_pre_acc_pos_violin}
    \vspace{-15pt}
\end{figure}

The Fig.~\ref{fig:sim_pre_acc_pos_violin} violin plots depict the distribution of position
precision for {\slamtoolbox} and the stereo SLAM methods.
For {\slamtoolbox} and {\gfgg}, map use concentrates the distribution at lower values.  
While {\slamtoolbox} has four times lower mean precision error (1.75 cm vs. 6.41 cm), both
have similar performance from a task perspective.


\subsubsection{Real-World}
Analysis will focus only on position precision, as the average orientation precision value
lies below 10 degrees for all methods.  Being at $1/8$ of the camera's field of view, this
level of orientation precision is acceptable from a task-driven perspective. 
Fig.~\ref{fig:realworld_pre_cum} provides cumulative position precision plots for the
real-world tests (w/o and w/map). 
{\hectorslam, {\dsol}, and {\msckf} did not perform well enough to warrant inclusion (below 35\%).
Performance of the other implementations roughly matches the ordering of simulation, with
{\slamtoolbox} and {\gfgg} achieving perfect completion at one robot diameter.
The LIO methods and {\orb} achieved high completion rates.
Passive stereo implementations continue to exhibit comparable performance to LiDAR-based
implementations.

Looking at the Fig.~\ref{fig:realworld_pre_cum} w/-map case, there is a left-ward shift in the curve
indicating improved precision and an overall compression in pose variance across methods.
The violin plots of Fig.~\ref{fig:realworld_pre_cum} (top), support this same finding.
{\slamtoolbox} and {\gfgg} manage $100\%$ completion (N-AUC: 0.93 vs. 0.90).
{\orb} and {\liosam} show a substantial decrease in completion and N-AUC. 
This decrease is due to failure during the initial mapping phase on the long-distance test. 
In the short-distance run, {\gfgg} and {\orb} performed comparably to {\slamtoolbox} and {\liosam}, achieving N-AUC scores of 0.96 and 0.97 versus 0.94 and 0.86, respectively.
{\gfgg} performs similarly across both testing modes, indicating that future improvements should target long-term mapping graph maintenance and map exploitation for localization in Visual SLAM.
The multimedia attachment provides visual evidence from the cameras to qualitatively see how
precision varies across methods.



\begin{figure}[t]
  \vspace{0.06in}
    \centering
    \begin{tikzpicture}[inner sep=0pt, outer sep=0pt]
        \node[anchor=north west] (VP) at (0,0) 
            {\includegraphics[width=0.48\textwidth]{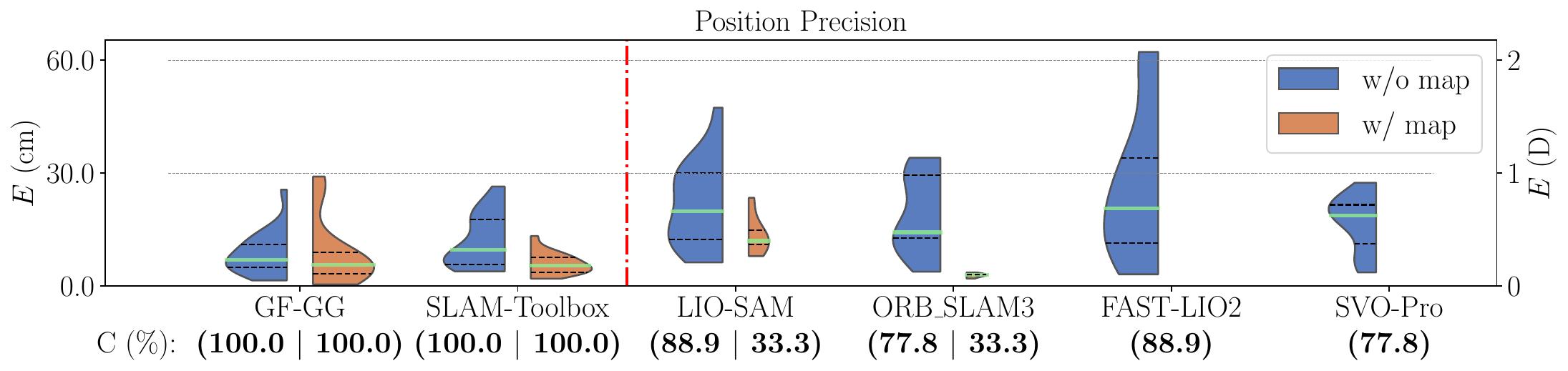}};
        \node[anchor=north,xshift=-35pt,yshift=-8pt] (CvsP_Legend) at ($(VP.south)+(3.0em, 0.0em)$)
            {\includegraphics[width=0.3\textwidth]{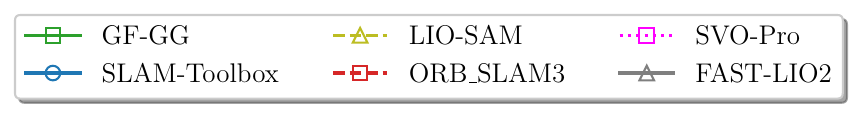}};
        \node[anchor=north,xshift=-35pt,yshift=-8pt] (CvsP_WO_Map) at ($(CvsP_Legend.south)+(-3.0em, 0.5em)$)
            {\includegraphics[width=0.18\textwidth]{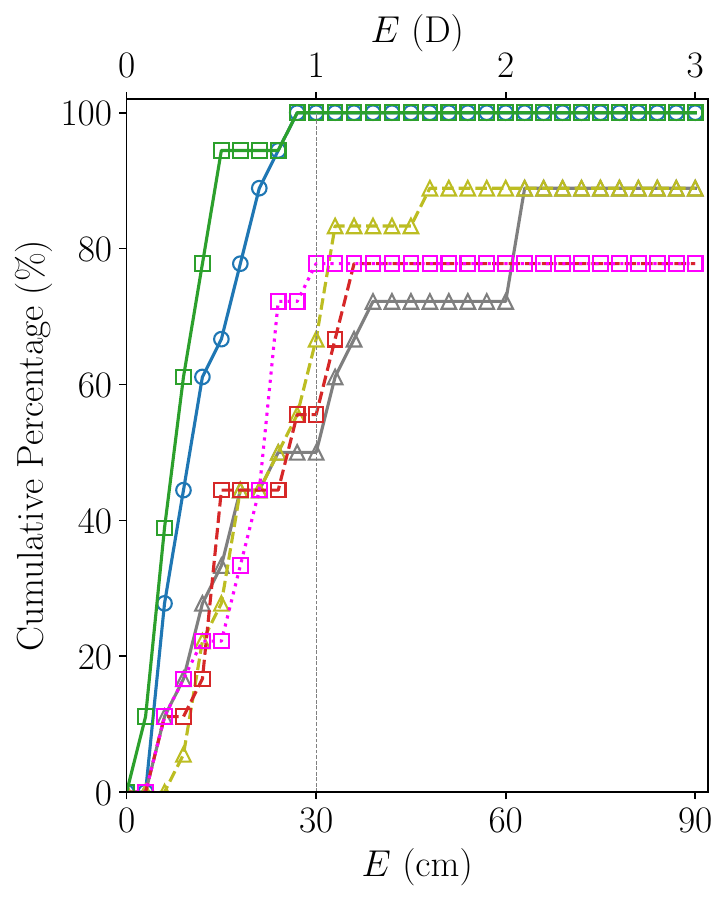}};
        \node[align=center, anchor=south] at ($(CvsP_WO_Map.south)+(1.0em, -0.8em)$) {\scriptsize w/o map };
        \node[anchor=west] (CvsP_W_Map) at (CvsP_WO_Map.east)
            {\includegraphics[clip=true,trim=1cm 0 0cm 0cm, width=0.16\textwidth]{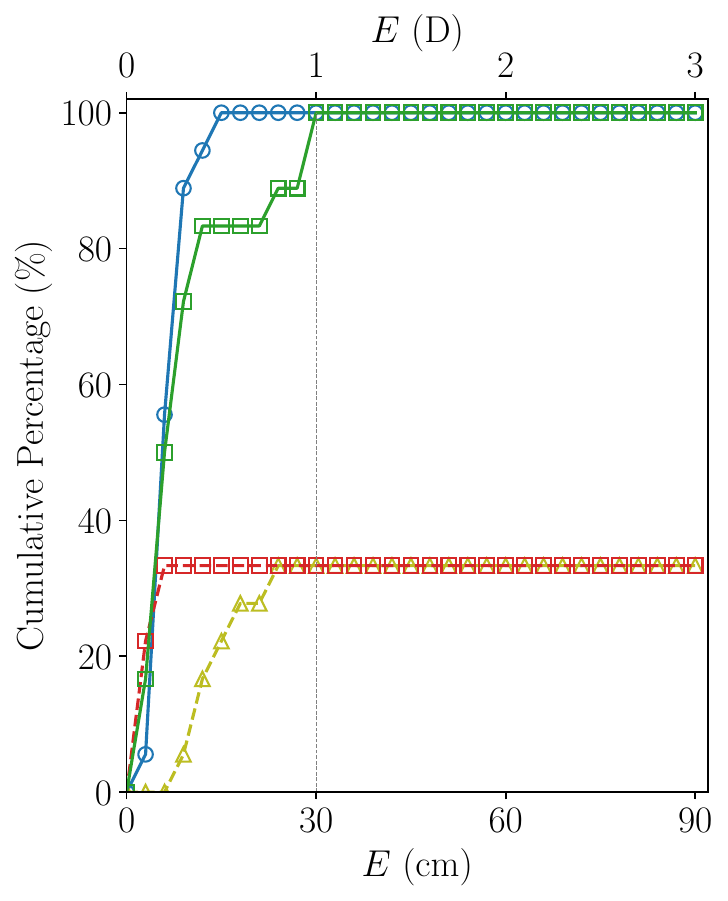}};
        \node[align=center, anchor=south] at ($(CvsP_W_Map.south)+(0.6em, -0.8em)$) {\scriptsize w/ map };
        \node[anchor=west] (CvsP_NUC) at ($(CvsP_W_Map.east)+(-0.em,-0.6em)$)
            {\includegraphics[width=0.12\textwidth]{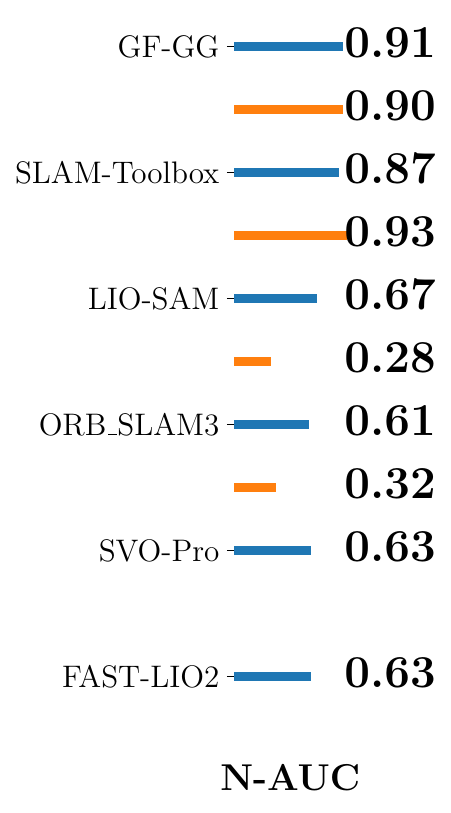}};

    \draw [dashed] ($(VP.south west)+(-0.0em, -0.5em)$) -- ($(VP.south east)+(-0.4em, -0.5em)$);

    \end{tikzpicture}
    \caption{
    Top: Distribution of Waypoint Position Precision for tests without (w/o) and with (w/) map in real-world. The legends and settings are consistent with Fig.~\ref{fig:sim_pre_acc_pos_violin}.
    Bottom: Cumulative Position Precision Plot of real-world test data with the same legends and settings as Fig.~\ref{fig:sim_wpt_pre_cum}.}
    \label{fig:realworld_pre_cum}
    \label{fig:realworld_pre_pos_violin}
    \label{fig:realworld_pre}
    \vspace{-15pt}
\end{figure}

\subsubsection{Discussion}
The \textit{TaskSLAM-Bench} results reveal that Visual SLAM achieves performance
approaching that of LiDAR SLAM for closed-loop navigation tasks, a finding not captured by
existing benchmarks.  These results demonstrate that vision-based approaches are capable of
delivering strong navigation performance once robust program execution is achieved.  Using a
SLAM map with map-to-frame tracking significantly improves performance. It reduces drift and
enhances long-term consistency, which has positive impacts on maintaining robust and
accurate navigation.

Long-path tests reveal that LiDAR SLAM scales more effectively to large environments and
handles feature-less scenarios better than visual methods. It excels in structured settings,
leveraging geometric features for robust loop closure and an efficient occupancy-grid map
representation.
Visual SLAM has issues with bounded map maintenance and feature poor areas
(e.g., long corridors), highlighting the need for multisensor fusion to enhance robustness.

Overall, {\gfgg} outperforms {\orb} by prioritizing efficient map-to-frame tracking and
local BA, suggesting scalability improvements are critical to long-term SLAM. Better
integrating robot odometry into the pose graph design can enhance robustness during 
feature poor scenarios to ensure continuous operation.

\section{Conclusion \label{sec:conclusion}}
This work introduces a task-driven SLAM benchmark, whose focus on precision highlights its
importance in task-oriented applications where repeatability is essential.
It considers the SLAM system mapping capabilities, an aspect often overlooked in existing
evaluations, but which provides a more comprehensive understanding of SLAM performance and
suitability for real-world tasks.
\textit{TaskSLAM-Bench} results show that visual SLAM methods can achieve precision
performance comparable to LiDAR-based methods in indoor navigation tasks. Further effort is
needed to resolve large-scale visual maps.  This finding underscores the potential of visual
SLAM systems to be reliable and effective alternatives to LiDAR approaches.  The benchmark
itself is low-cost and easy to implement. Other researchers may test their robots in their
own environments with a given SLAM implementation. 

\balance

\bibliographystyle{IEEEtran}
\bibliography{bibliography}

\end{document}